\documentclass[letterpaper]{article} % DO NOT CHANGE THIS
\usepackage{aaai25}  % DO NOT CHANGE THIS
\usepackage{times}  % DO NOT CHANGE THIS
\usepackage{helvet}  % DO NOT CHANGE THIS
\usepackage{courier}  % DO NOT CHANGE THIS
\usepackage[hyphens]{url}  % DO NOT CHANGE THIS
\usepackage{graphicx} % DO NOT CHANGE THIS
\usepackage{natbib}  % DO NOT CHANGE THIS AND DO NOT ADD ANY OPTIONS TO IT
\usepackage{caption} % DO NOT CHANGE THIS AND DO NOT ADD ANY OPTIONS TO IT
\usepackage{xcolor}
\usepackage{enumitem}
\usepackage{algorithm}
\usepackage{algorithmic}

\usepackage{newfloat}
\usepackage{listings}
\DeclareCaptionStyle{ruled}{labelfont=normalfont,labelsep=colon,strut=off} % DO NOT CHANGE THIS
\floatstyle{ruled}
\newfloat{listing}{tb}{lst}{}
\floatname{listing}{Listing}
\title{A Socratic RAG Approach to Connect Natural Language Queries on Research Topics with Knowledge Organization Systems}
\author{
   Lew Lefton\textsuperscript{\rm 1},\
   Kexin Rong\textsuperscript{\rm 1},\
   Chinar Dankhara\textsuperscript{\rm 1},\
   Lila Ghemri\textsuperscript{\rm 2},\
   Firdous Kausar\textsuperscript{\rm 3},\
   A. Hannibal Hamdallahi\textsuperscript{\rm 3}
}
\affiliations{
\textsuperscript{\rm 1}Georgia Institute of Technology\\
\textsuperscript{\rm 2}Texas Southern University\\
\textsuperscript{\rm 3}Fisk University\\
lew.lefton@gatech.edu, krong@gatech.edu, chinardankhara@gatech.edu, lila.ghemri@tsu.edu, fkausar@fisk.edu, aleach@fisk.edu
}

\begin{document}

\maketitle{}

\begin{abstract}
In this paper, we propose a Retrieval Augmented Generation (RAG) agent that maps natural language queries about research topics to precise, machine-interpretable semantic entities. Our approach combines RAG with Socratic dialogue to align users' intuitive understanding of research topics with established Knowledge Organization Systems (KOSs). The proposed approach will effectively bridge "little semantics" (domain-specific KOS structures) with "big semantics" (broad bibliometric repositories), making complex academic taxonomies more accessible. Such agents have the potential for broad use. We illustrate with a sample application called CollabNext, which is a person-centric knowledge graph connecting people, organizations, and research topics. We further describe how the application design has an intentional focus on HBCUs and emerging researchers to raise visibility of people historically rendered invisible in the current science system.

\end{abstract}


\section{Introduction}
What is a research topic? Research is carried out on almost everything, so the concept space is wide and deep. It is unsurprising that there is no universal resource that covers all research concepts. Instead, many domain-specific Knowledge Organization Systems (KOS) have emerged providing ontologies, taxonomies, lists, and thesauri of their areas \cite{salatino:2024}. These KOSs are good sources of expert-curated ground truth. Unfortunately, in a KOS, topic data typically consists only of strings and not semantic entities, which makes research topics a weak link in knowledge axiomatization and research classification.

A knee-jerk reaction to this challenge is to leverage a Large Language Model (LLM) to generate research topics, based on a corpus of research papers, datasets, etc. In fact, there are many tools in the area of topic classification that do precisely that. Unfortunately, LLM generated topics do not align well with how human beings talk and think about research. 

Alignment of natural language research topics with structured KOSs is needed for CollabNext \cite{CollabNext}, an application currently under development as part of the National Science Foundation's Prototype Open Knowledge Network \cite{ProtoOKN}.  CollabNext allows users to explore who is working in what research area and where they are. While this seems simple and straightforward to answer, it turns out to be quite nuanced and challenging. Name collisions make it difficult to identify the "who", especially for common names or Asian names \cite{namedisambig}. Organizations, the "where", can also have multiple names, campuses, etc.  Fortunately, for both individuals and organizations, there are established and open semantic structures, for example in schema.org \cite{schemaorg} or Wikidata.org \cite{wikidata}, and available persistent identifiers, e.g. ORCID \cite{orcid} and ROR \cite{ror}.
Research topics, the "what", turn out to be more complicated. There is no universal list of research concepts, and certainly nothing that compares to a persistent identifier. 

This paper begins with a brief overview of related work on the challenge of organizing research topics. We then provide a short description of CollabNext, including its intentional design to increase the visibility of emerging researchers. We share an example that highlights the challenge and importance of aligning natural language research topic queries with robust, open, and structured data in KOSs. Next we proceed to lay out a plan for a RAG agent designed to meet this challenge using a novel Socratic approach. The final section has a conclusion and considers the next steps.

\section{Related Work}
In this section, we review approaches for automatically generating topics in scientific publications. We focus on three main paradigms: supervised topic classification, unsupervised topic modeling, and knowledge organization systems.

\subsection{Topic Classification} Topic classification takes a supervised approach, using often manually curated document topics to train models for categorizing new publications. However, obtaining high-quality ground truth labels at scale remains challenging, forcing many methods to build topic taxonomies from scratch~\cite{liu2014hierarchical}.

The Microsoft Academic Graph (MAG)~\cite{sinha2015overview} represents a significant effort in this direction. 
Their methodology began with manual definition of a small number of top-level concepts, followed by extraction of 28,000 academic concepts from Wikipedia articles~\cite{shen2018web}. MAG implemented topic assignment as a multilabel classification problem, enabling papers to be tagged with multiple concepts.
After the discontinuation of MAG in 2021, OpenAlex attempted to reproduce the topic classification model using historical MAG data as training labels~\cite{priem2022openalex}. During this process, OpenAlex identified several limitations of MAG concepts, including term polysemy, ambiguity, and the static nature of concepts that failed to evolve with research trends.
OpenAlex subsequently adopted a labeled dataset from CWTS~\cite{van2024open} and integrated it with concepts from the All Science Journal Classification (ASJC) codes. 
However, obtaining high-quality training labels at scale remains a challenge, often forcing methods to construct topic taxonomies from scratch~\cite{liu2014hierarchical}.

\subsection{Topic Modeling} 
Unsupervised topic modeling techniques discover common themes within document collections by clustering similar documents and deriving topic labels from shared characteristics within clusters.
For scientific papers, these methods typically leverage multiple signals including semantic similarity from titles and abstracts, citation network structures, and shared references.
Domain-specific pretrained models like SciBERT~\cite{hosokawa2024} have been shown to enhance clustering quality of abstract embeddings by using purpose-built tokenizers and specialized pretraining that capture field-specific terminology.

Similarly, BERTopic~\cite{grootendorst2022bertopic} uses a pre-trained language model to generate document embedding, performs dimensionality reduction, and creates semantically similar clusters (using HDBSCAN) that each represent a distinct topic. BERTopic uses a variant of TF-IDF to extract topic representations from each topic. 
CWTS~\cite{van2024open} leverages large language models (LLMs) to generate research topic labels. The LLM receives titles of the 250 most cited publications in each cluster and generates a label and a brief summary of the research area. 

However, clustering-based approaches face limitations: extracted topics often fail to align with established ontologies, and poor clustering quality can result in topics lacking clear interpretability.

\subsection{Knowledge Organization Systems} 

A Knowledge Organization System (KOS) \cite{salatino:2024} is a structured framework that enables the arrangement, management, retrieval, and dissemination of information. 
These systems are crucial to library science, information retrieval, and knowledge management, offering standardized vocabularies and defined relationships between concepts to improve information accessibility and interoperability~\cite{Hodge:2000}. 

KOS encompasses several key methodologies:
\begin{itemize}[leftmargin=*]
    \item \emph{Classification schemes}: Systems such as the Dewey Decimal Classification (DDC)~\cite{Dewey:2011} and Library of Congress Classification (LCC) systematically group related concepts based on shared characteristics, providing logical organization for library resources.
    \item \emph{Thesauri}: Controlled vocabularies that map relationships between terms, including synonyms, antonyms, and hierarchical relationships. A prominent example is the Medical Subject Headings (MeSH), which is the controlled vocabulary used to index the huge amount of biomedical literature \cite{NLM:2022}.
    \item \emph{Taxonomies}: Hierarchical systems organizing concepts into parent-child relationships, commonly used in web navigation and content management systems.
    \item \emph{Ontologies}: Formal and machine-interpretable specifications that define domain concepts and their relationships~\cite{Guarino:2009}. These are crucial for semantic web technologies and AI applications, enabling data interoperability and automated reasoning.
    \item \emph{Controlled Vocabularies}: Curated term lists ensure consistent language across collections, enhancing retrieval precision and accuracy.
\end{itemize}

KOSs help to uncover knowledge from various sources, especially in the context of big data and the semantic web. Integrating and discovering knowledge from heterogeneous sources becomes a lot easier with KOS, which is also very usefu; for advanced research and development tasks. The Simple Knowledge Organization System (SKOS) presents a model for sharing and linking knowledge organization systems on the web \cite{Miles:2009}. This W3C-endorsed model based on Resource Description Framework (RDF) promotes interoperability, as well as the seamless exchange of information between heterogeneous systems. 
%%\textcolor{blue}{KR: Is SKOS important to cover?}
%% Yes, it was mentioned several times at the ISWC so I believe it is still an active w3c standard https://www.w3.org/TR/skos-reference/

Despite their utility, KOSs encounter some issues, such as how to remain pertinent in rapidly changing areas of knowledge and how to ensure that they work well with other systems. \cite{lauruhn:2016} comment on the need for KOSs to have adaptive management strategies that allow them to change along with our knowledge structures.

Recent research shows promises of integrating KOS with large language models (LLMs) to enhance AI explainability. \cite{ahmed:2023} demonstrated that the combination of knowledge graphs with LLM produces more interpretable outputs than LLMs alone, while \cite{krause:2024} showed that LLMs coupled with external knowledge sources improve commonsense reasoning capabilities in AI systems.

\section{CollabNext}

There are many potentially impactful applications which would benefit from having a RAG agent that provides an explainable topic identifier based on established ground truth of human-generated KOSs. To serve as an illustrative example, we describe CollabNext \cite{CollabNext} which is currently under development as part of the NSF's Prototype Open Knowledge Network \cite{ProtoOKN}. This application depends on open, robust, and structured data on research topics.

\subsection{Application Overview}
CollabNext implements a knowledge graph with entities consisting of people, organizations, and research topics. The primary data source is currently the bibliometric dataset from OpenAlex \cite{priem2022openalex}, but the intention is to include other relevant data sources as development continues. Relationships between people and organizations are available directly in OpenAlex via author and institution tables. Relationships between people and topics are inferred, since people are connected to OpenAlex works as authors, and works are connected to topics. See Figure 1 for a conceptual schema of the CollabNext knowledge graph. Note, that the schema in Figure 1 includes additional data like Grants and Patents which are not yet in OpenAlex, but these objects could still be assigned topics via methods discussed above.

\begin{figure}[t]
\centering
\includegraphics[width=0.9\columnwidth]{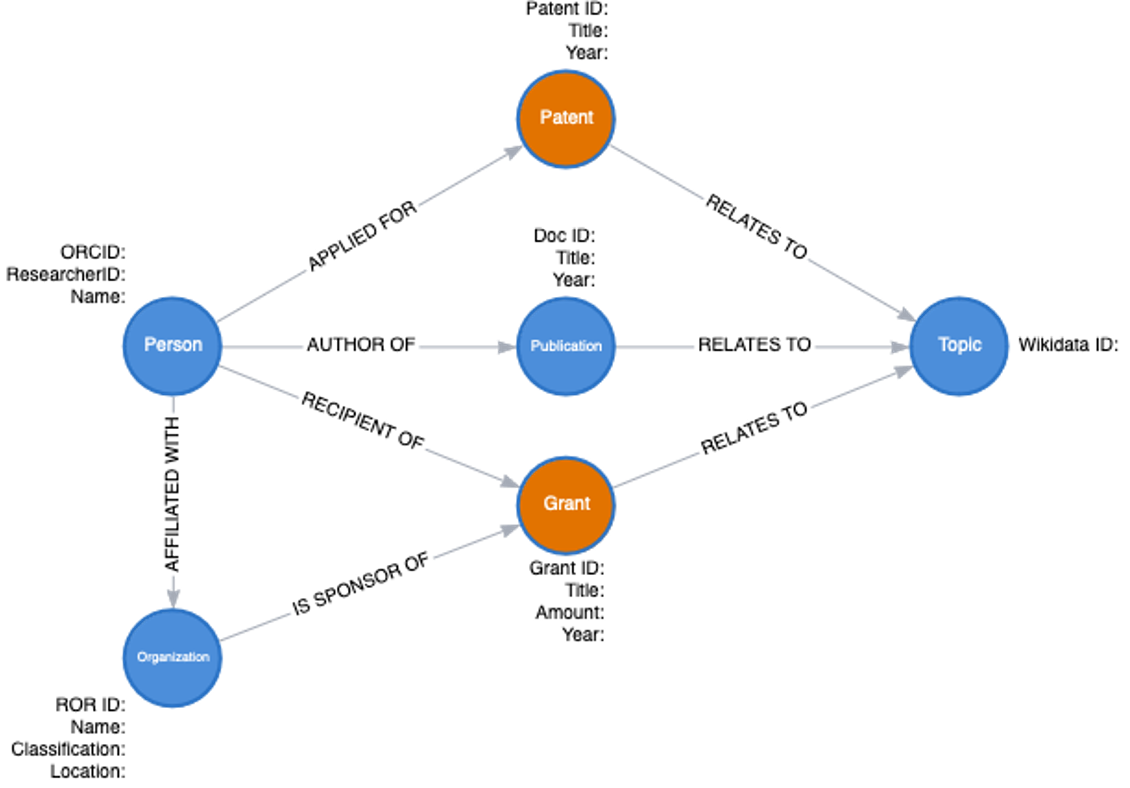} % Reduce the figure size so that it is slightly narrower than the column. Don't use precise values for figure width.This setup will avoid overfull boxes.
\caption{Conceptual schema of CollabNext knowledge graph}
\label{fig1}
\end{figure}

The CollabNext application is being developed using an intentional design approach, initially prioritizing Historically Black Colleges and Universities (HBCUs) and emerging researchers. This is a deliberate effort to counterbalance the accumulated advantage of well-resourced research organizations, which is an example of the Matthew Effect \cite{Merton-Matthew-1968}. This effect is a common phenomenon in social systems, and can be summarized as "the rich get richer and the poor get poorer." CollabNext is being designed to counterbalance this bias, specifically in the social system of science research \cite{Bol2018, Rossiter1993, Petersen2011}.

One of the ways to do this is to, by default, focus on HBCUs and other organization where there are a high percentage of underrepresented researchers. Another design decision which can help raise visibility of invisible researchers is to not order researchers by citation count, which exacerbates the Matthew effect, but rather default to use geospatial data like location and distance as a primary filter.  This is effectively adding an implicit "near me" search filter and showing people who work on a specific research topic who are also geographically close, e.g. at your same institution, in your same state, or your same timezone. The rationale for this approach is based on work in Team Science, which shows that geographical distance is a major barrier for scientific collaboration \cite{HOEKMAN2024104927}. The work of \cite{Katz1994-rv} shows that the probability of collaboration decreases as distance grows, based on co-authorship relations.

\subsection{Challenge: Topic Search Needs an NLP interface}
During the development of CollabNext, it became apparent that there was a disconnect between how people see research topics and how research topics are classified by AIs.  We will illustrate with an example.

Suppose a user wants to find researchers who are working on the topic of \textit{plastic recycling} at a nearby University.  There is no OpenAlex topic that directly matches this.  Matching just for \textit{plastic} yields 14 topic ids in the subfield  \textit{Polymers and Plastics}. This topic could also be considered \textit{Waste Management and Disposal} which is in the subfield \textit{Environmental Science}. Or it may be the topic \textit{Biodegradable Polymers as Biomaterials and Packaging} which is in the subfield \textit{Biomaterials}.  It may even be the case that the user is not interested in the Chemistry of plastic recycling at all, but rather the economics or logistics of it. Similar collisions and ambiguities arise when the keyword \textit{recycling} is used. More clarification is needed from the user. This subtle but important challenge can be improved with the RAG agent described below. 

\section{Proposed Approach}
%\textcolor{blue}{KR: Wrote a version based on my understanding. Feel free to edit. }
% Looks great! Thank you.

We propose an LLM agent leveraging multi-round Retrieval-Augmented Generation (RAG) to bridge the semantic gap between users' natural language queries on research topics and structured KOSs. While LLMs have demonstrated remarkable capabilities, they often struggle with domain-specific knowledge and are prone to hallucinations~\cite{zhao2023survey}, particularly in specialized academic contexts. Retrieval-augmented approaches have emerged as a promising solution by grounding LLM outputs in external knowledge bases~\cite{lewis2020retrieval}.

Traditional RAG systems typically employ single-round retrieval, where knowledge is retrieved solely based on the initial query~\cite{izacard2022few}.  However, this approach proves insufficient for complex questions that require iterative refinement and disambiguation, such as mapping colloquial research descriptions to formal taxonomic structures. 
Recent work has demonstrated the effectiveness of multi-round retrieval and reasoning~\cite{borgeaud2022improving,shao2023enhancing,zhuang2024efficientrag}, where each round of interaction refines the query and retrieval process. 

Building on these advances, we propose a Socratic dialogue approach where the agent guides users through structured exploration of research topics, by asking questions of users, rather than simply offering answers. We anticipate that this dialog will be more than simple clarification questions. The agent should engage in meaningful conversation to examine and understand the user's research topic of interest at multiple levels of granularity. The agent will be guided by reasoning across the multiple KOS's on which it was trained. By grounding these conversations in human-curated KOSs as the knowledge foundation, the system provides transparent data provenance and explainable recommendations.

\begin{figure}[t]
\centering
\includegraphics[width=0.95\columnwidth]{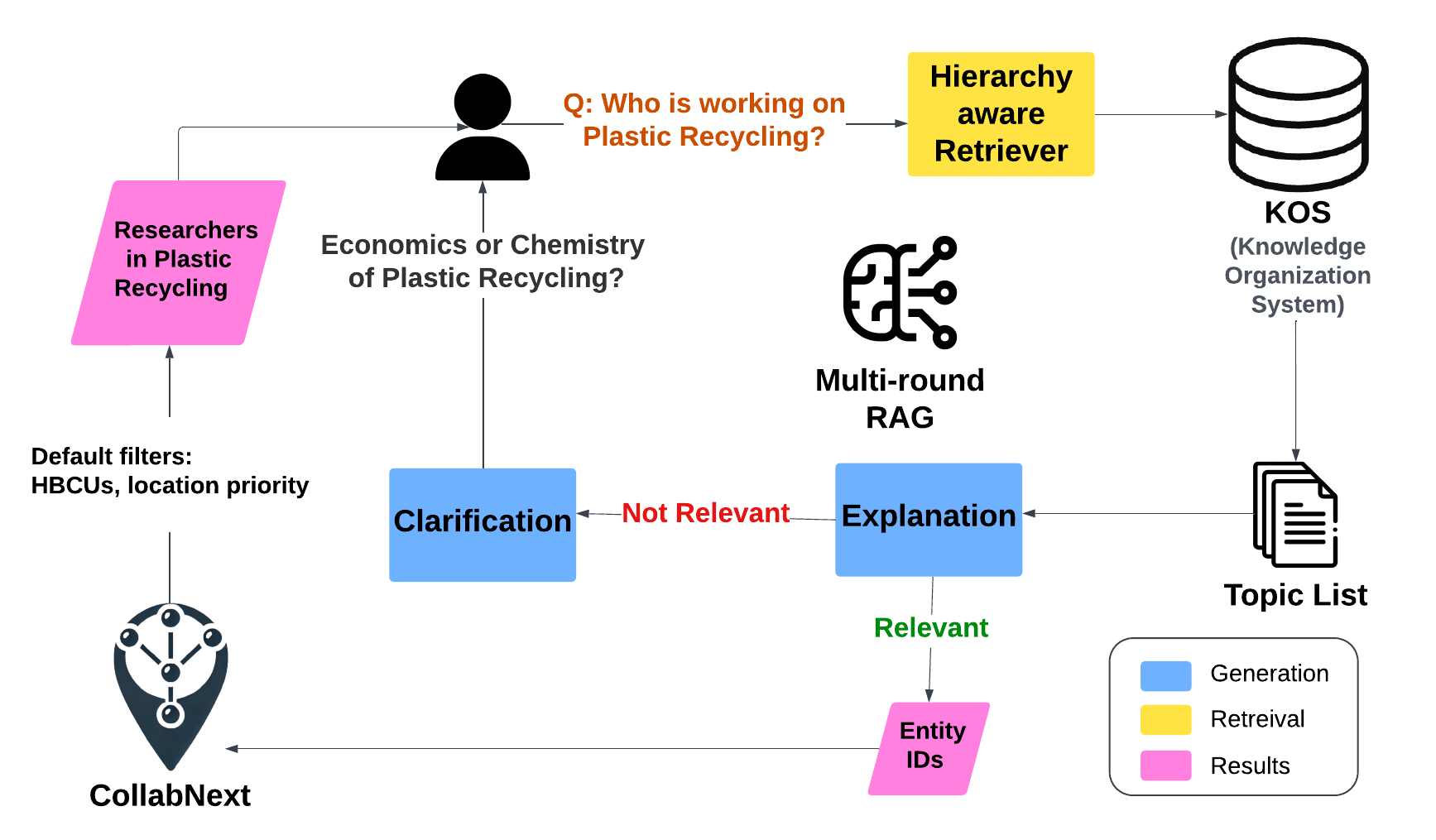} % Reduce the figure size so that it is slightly narrower than the column. Don't use precise values for figure width.This setup will avoid overfull boxes.
\caption{Proposed multi-round RAG design that bridges natural language queries on research topics with structured KOS data. }
\label{fig2}
\end{figure}

If successful, the proposed agent would be able to effectively bridge "little semantics" (domain-specific KOS structures) with "big semantics" (broad bibliometric repositories), making complex academic taxonomies more accessible to users who naturally express research interests in colloquial language. This approach enables precise mapping between informal research descriptions and formal knowledge structures while maintaining transparency and interpretability. Figure~\ref{fig2} illustrates the high-level workflow of our proposed system, and we overview the detailed design below.

\subsubsection{Topic Retrieval}
To effectively identify relevant research topics from KOS taxonomies, we propose a novel hierarchical retrieval mechanism. We will use domain-specific, pretrained models such as SciBERT~\cite{beltagy2019scibert} to generate embeddings for each topic node. The retrieval process occurs in two stages:
\begin{enumerate}
    \item Initial Semantic Search: Topic embeddings are represented as dense embeddings of topic title and description strings. These are retrieved based on embedding similarity with the user query, using cosine similarity normalized by a temperature parameter $\tau$ to control retrieval diversity - higher values generate more diverse candidates while lower values prioritize closer semantic matches.
    \item Hierarchy-Aware Reranking: Given the initial topic list, we rerank them by considering the ontological structure of topics. The reranking score combines three components: (1) direct semantic similarity between query and topic; (2) weighted ancestor path similarity computed as $$score(t)= sim(q,t)+\alpha\sum_{a \in ancestors(t)} \beta^d \times sim(q, a)$$ where $\alpha$ controls the overall influence of ancestral relationships and $\beta$ determines the decay rate with ancestral distance $d$; and (3) sibling coherence that boosts topics whose siblings also show high query similarity. This scoring function ensures that retrieved topics are both semantically relevant and structurally coherent within their respective taxonomies.
\end{enumerate}

\subsubsection{Multi-round RAG}
The academic knowledge landscape is characterized by fragmented KOS resources: broad multi-field systems that cover multiple disciplines but lack depth, and specialized single-field KOS that provide granular topic categorization within specific domains. Our proposed multi-round RAG process is specifically designed to navigate this fragmented structure through a two-phase approach.

In the first phase, the system queries broad multi-field KOS systems such as All Science Journal Classification Codes, OpenAIRE’s Field of Science Taxonomy, or Dewey Decimal Classification to identify relevant high-level research areas. For each retrieved topic, the LLM generates contextual explanations by synthesizing definitions and descriptions from linked knowledge bases. The agent engages in iterative dialogue with users, presenting candidate topics along with their explanations and soliciting feedback to refine the search. This phase continues until users confirm their broad research direction.

The second phase starts when users confirm topics that have corresponding single-field KOS coverage. For instance, in Computer Science, the system can leverage multiple specialized taxonomies including the Computer Science Ontology (CSO), ACM Computing Classification System (CCS), and Wikipedia's Computer Science Subject Headings. The agent uses context accumulated from the first phase to guide exploration within these detailed taxonomies. For fields lacking comprehensive KOS coverage~\cite{salatino:2024}, the system employs more stringent filtering within multi-field KOS to identify the most precise available classifications.

\section{Conclusion and Next Steps}

We have outlined a RAG agent that takes natural language queries about research topics and provides focused machine-representable semantic entity identifiers based on the ground truth of human generated KOSs. We have also described the CollabNext application. One of the goals of CollabNext is to construct a knowledge graph that makes visible the relationships between people, research organizations, and research topics. CollabNext is being intentionally designed to make emerging researchers more visible and thereby help reduce the Matthew Effect in scientific research. We  
further illustrate how the RAG agent could be used, and provided an approach to building the agent.

The agent described here has potential beyond the use case of CollabNext. Indeed, \textit{any application that links its data to research topics via an existing KOS could also leverage the agent as a research topic "Rosetta Stone."} Moreover, disparate research artifacts could be semantically linked using research topics as an index, including articles, datasets, websites, AI models, GitHub repositories, theses, posters, etc. An existing example of this is the OpenAlex "aboutness" endpoint. This could lead to broader integration of research knowledge and data across diverse domains using a well-defined topic architecture. New research topics at disciplinary boundaries could also emerge and be naturally included as domain specific KOSs incorporate new structure and data. 

We note that there are multiple other data-driven products which are relevant for this discussion.   Examples include Google Scholar, LinkedIn (owned by Microsoft), Clarivate, Elsevier, Semantic Scholar, Metaphacts, System.com, etc. Some of them are commercial and are only available to well resourced institutions. Others leverage proprietary data and are not well suited to open systems. Also, these tools are more focused on research artifacts, and less on the researchers themselves.

In some sense, the RAG agent serves the role of a topic disambiguation tool.  This is clearly needed for tools that work at the interface of humans and AIs.  Nuance and subtlety of how different researchers think about the “same” thing may provide insight and new results, and allow researchers to see integrative work and connect with adjacent researchers whom they did now know about.

This paper is admittedly only a high level description of the agent. We are taking to heart the workshop submission guidelines \cite{tika25} which welcome short papers \textit{focused on case studies, work-in-progress, or visionary ideas}. There are many unanswered questions including
\begin{enumerate}
    \item What happens if there is a topic that cannot be resolved by the KOSs that the agent knows about. This knowledge gap could be a signal of an emerging research area or perhaps a need for KOS improvement. Some KOSs are updated more frequently than others, so this may be an opportunity to provide a feedback loop.
    \item Would it be possible for users to explore topics both up and down a refinement, i.e. broadening and narrowing vertically, as well as exploring adjacent siblings (horizontally). 
    \item How would user interactions be incorporated into the agent's knowledge over time? Could histories and usage patterns help with noisy/overlapping topics, or identify frequently asked questions that could indicate gaps between user needs and KOS data structures.
\end{enumerate}

The CollabNext team has plans to build a simple proof-of-concept prototype of the proposed RAG agent. Such a prototype would likely be focused on only a handful of available KOSs, but would likely include both OpenAlex and Wikidata as a starting point. 

\section{Acknowledgments}
This work was supported by the National Science Foundation TIP Directorate under Award Number 2333737. We would also like to acknowledge helpful guidance from the reviewers of this submission, along with the following members of the CollabNext Team who have helped develop the ideas here and the alpha version of the CollabNext application: Didier Contis, Kinnis Gosha, John Porter, Christopher Thomas, Craig Abbey, Leslie Collins, Sufyan Baksh, Sajid Hussain, Robert Briggs, Samarth Chandna, Sambridhi Deo, Lazarus Egwurube, Diamond GC, Ilia Khalighi, Vidushi Maheshwari, Ikechukwu Mgbemele, Christian Moore, Rejin Nepal, Mingxuan Nie, Netra Nyaupane, Nirmal Patel, Maruti Ram Ponnaganti, Ram Sharma, Keller Smith, Siddharth Singh Solanki, and Jaelyn Sykes.

\bibliography{aaai25}

\begin{thebibliography}{37}
\providecommand{\natexlab}[1]{#1}

\bibitem[{AAAI(2025)}]{tika25}
AAAI. 2025.
\newblock Workshop on A Translational Institute for Knowledge Axiomatization.
\newblock Website.

\bibitem[{Ahmed et~al.(2023)Ahmed, Firmansyah, Sherif, Moussallem, and Ngonga~Ngomo}]{ahmed:2023}
Ahmed, A.~F.; Firmansyah, A.~F.; Sherif, M.~A.; Moussallem, D.; and Ngonga~Ngomo, A.-C. 2023.
\newblock Explainable Integration of Knowledge Graphs Using Large Language Models.
\newblock In \emph{Natural Language Processing and Information Systems}, 124--139. Springer.

\bibitem[{Beltagy, Lo, and Cohan(2019)}]{beltagy2019scibert}
Beltagy, I.; Lo, K.; and Cohan, A. 2019.
\newblock SciBERT: A Pretrained Language Model for Scientific Text.
\newblock In \emph{Proceedings of the 2019 Conference on Empirical Methods in Natural Language Processing}, 3615--3620.

\bibitem[{Bol, de~Vaan, and van~de Rijt(2018)}]{Bol2018}
Bol, T.; de~Vaan, M.; and van~de Rijt, A. 2018.
\newblock The Matthew effect in science funding.
\newblock \emph{Proceedings of the National Academy of Sciences}, 115(19): 4887--4890.

\bibitem[{Borgeaud et~al.(2022)Borgeaud, Mensch, Hoffmann, Cai, Rutherford, Millican, Van Den~Driessche, Lespiau, Damoc, Clark et~al.}]{borgeaud2022improving}
Borgeaud, S.; Mensch, A.; Hoffmann, J.; Cai, T.; Rutherford, E.; Millican, K.; Van Den~Driessche, G.~B.; Lespiau, J.-B.; Damoc, B.; Clark, A.; et~al. 2022.
\newblock Improving language models by retrieving from trillions of tokens.
\newblock In \emph{International conference on machine learning}, 2206--2240. PMLR.

\bibitem[{CollabNext(2024)}]{CollabNext}
CollabNext. 2024.
\newblock CollabNext.io.
\newblock Website.

\bibitem[{Dewey et~al.(2011)Dewey, Mitchell, Beall, Martin, Matthews, and New}]{Dewey:2011}
Dewey, M.; Mitchell, J.~S.; Beall, J.; Martin, G.; Matthews, W.~E.; and New, G.~R. 2011.
\newblock \emph{Dewey Decimal Classification and Relative Index}.
\newblock Dublin, Ohio: OCLC Online Computer Library Center, 23rd edition.
\newblock ISBN 978-0910608879.

\bibitem[{Grootendorst(2022)}]{grootendorst2022bertopic}
Grootendorst, M. 2022.
\newblock BERTopic: Neural topic modeling with a class-based TF-IDF procedure.
\newblock \emph{arXiv preprint arXiv:2203.05794}.

\bibitem[{Guarino, Oberle, and Staab(2009)}]{Guarino:2009}
Guarino, N.; Oberle, D.; and Staab, S. 2009.
\newblock What Is an Ontology?
\newblock In Staab, S.; and Studer, R., eds., \emph{Handbook on Ontologies}, 1--17. Berlin, Germany: Springer, 2nd edition.

\bibitem[{Hodge(2000)}]{Hodge:2000}
Hodge, G. 2000.
\newblock Systems of Knowledge Organization for Digital Libraries: Beyond Traditional Authority Files.
\newblock Technical Report Report 91, Council on Library and Information Resources, Washington, D.C., USA.

\bibitem[{Hoekman and Rake(2024)}]{HOEKMAN2024104927}
Hoekman, J.; and Rake, B. 2024.
\newblock Geography of authorship: How geography shapes authorship attribution in big team science.
\newblock \emph{Research Policy}, 53(2): 104927.

\bibitem[{Hosokawa et~al.(2024)Hosokawa, Yamato, Higashinaka, Kikui, and Sugiyama}]{hosokawa2024}
Hosokawa, R.; Yamato, J.; Higashinaka, R.; Kikui, G.; and Sugiyama, H. 2024.
\newblock Reference Classification Using BERT Models to Support Scientific-Document Writing.
\newblock In \emph{New Frontiers in Artificial Intelligence: JSAI-IsAI 2023 International Workshops}, 167--183. Springer.

\bibitem[{Izacard et~al.(2022)Izacard, Lewis, Lomeli, Hosseini, Petroni, Schick, Dwivedi-Yu, Joulin, Riedel, and Grave}]{izacard2022few}
Izacard, G.; Lewis, P.; Lomeli, M.; Hosseini, L.; Petroni, F.; Schick, T.; Dwivedi-Yu, J.; Joulin, A.; Riedel, S.; and Grave, E. 2022.
\newblock Few-shot learning with retrieval augmented language models.
\newblock \emph{arXiv preprint arXiv:2208.03299}, 2(3).

\bibitem[{Katz(1994)}]{Katz1994-rv}
Katz, J.~S. 1994.
\newblock Geographical proximity and scientific collaboration.
\newblock \emph{Scientometrics}, 31(1): 31--43.

\bibitem[{Kim, Kim, and Kim(2023)}]{namedisambig}
Kim, J.; Kim, J.; and Kim, J. 2023.
\newblock Effect of Chinese characters on machine learning for Chinese author name disambiguation: A counterfactual evaluation.
\newblock \emph{Journal of Information Science}, 49(3): 711--725.

\bibitem[{Krause and Stolzenburg(2024)}]{krause:2024}
Krause, S.; and Stolzenburg, F. 2024.
\newblock From Data to Commonsense Reasoning: The Use of Large Language Models for Explainable AI.
\newblock \emph{arXiv preprint arXiv:2407.03778}.

\bibitem[{Lauruhn and Groth(2016)}]{lauruhn:2016}
Lauruhn, M.; and Groth, P. 2016.
\newblock Sources of Change for Modern Knowledge Organization Systems.
\newblock \emph{Knowledge Organization}, 43(8): 622--629.

\bibitem[{Lewis et~al.(2020)Lewis, Perez, Piktus, Petroni, Karpukhin, Goyal, K{\"u}ttler, Lewis, Yih, Rockt{\"a}schel et~al.}]{lewis2020retrieval}
Lewis, P.; Perez, E.; Piktus, A.; Petroni, F.; Karpukhin, V.; Goyal, N.; K{\"u}ttler, H.; Lewis, M.; Yih, W.-t.; Rockt{\"a}schel, T.; et~al. 2020.
\newblock Retrieval-augmented generation for knowledge-intensive nlp tasks.
\newblock \emph{Advances in Neural Information Processing Systems}, 33: 9459--9474.

\bibitem[{Liu, Zhang, and Chen(2014)}]{liu2014hierarchical}
Liu, T.; Zhang, N.~L.; and Chen, P. 2014.
\newblock Hierarchical latent tree analysis for topic detection.
\newblock In \emph{Machine Learning and Knowledge Discovery in Databases: European Conference, ECML PKDD 2014, Nancy, France, September 15-19, 2014. Proceedings, Part II 14}, 256--272. Springer.

\bibitem[{Merton(1968)}]{Merton-Matthew-1968}
Merton, R.~K. 1968.
\newblock The Matthew effect in science.
\newblock \emph{Science}, 159(3810): 56--63.

\bibitem[{Miles and Bechhofer(2009)}]{Miles:2009}
Miles, A.; and Bechhofer, S. 2009.
\newblock SKOS Simple Knowledge Organization System Reference.
\newblock W3c recommendation, World Wide Web Consortium.

\bibitem[{NLM(2022)}]{NLM:2022}
NLM. 2022.
\newblock \emph{National Library of Medicine Medical Subject Headings (MeSH)}.
\newblock Bethesda, MD, USA.

\bibitem[{NSF(2023)}]{ProtoOKN}
NSF. 2023.
\newblock Prototype Open Knowledge Network.
\newblock Website.

\bibitem[{{ORCID Inc.}(2024)}]{orcid}
{ORCID Inc.} 2024.
\newblock {ORCID: Connecting Research and Researchers}.
\newblock \url{https://orcid.org}.
\newblock Accessed: 2024-11-24.

\bibitem[{Petersen et~al.(2011)Petersen, Jung, Yang, and Stanley}]{Petersen2011}
Petersen, A.~M.; Jung, W.-S.; Yang, J.-S.; and Stanley, H.~E. 2011.
\newblock Quantitative and empirical demonstration of the Matthew effect in a study of career longevity.
\newblock \emph{Proceedings of the National Academy of Sciences}, 108(1): 18--23.

\bibitem[{Priem, Piwowar, and Orr(2022)}]{priem2022openalex}
Priem, J.; Piwowar, H.; and Orr, R. 2022.
\newblock OpenAlex: A fully-open index of scholarly works, authors, venues, institutions, and concepts.
\newblock \emph{arXiv preprint arXiv:2205.01833}.

\bibitem[{{Research Organization Registry}(2024)}]{ror}
{Research Organization Registry}. 2024.
\newblock {Research Organization Registry (ROR)}.
\newblock \url{https://ror.org}.
\newblock Accessed: 2024-11-24.

\bibitem[{Rossiter(1993)}]{Rossiter1993}
Rossiter, M.~W. 1993.
\newblock The Matthew Matilda effect in science.
\newblock \emph{Social Studies of Science}, 23(2): 325--341.

\bibitem[{Salatino et~al.(2024)Salatino, Aggarwal, Mannocci, Osborne, and Motta}]{salatino:2024}
Salatino, A.; Aggarwal, T.; Mannocci, A.; Osborne, F.; and Motta, E. 2024.
\newblock A Survey on Knowledge Organization Systems of Research Fields: Resources and Challenges.
\newblock arXiv:2409.04432.

\bibitem[{Shao et~al.(2023)Shao, Gong, Shen, Huang, Duan, and Chen}]{shao2023enhancing}
Shao, Z.; Gong, Y.; Shen, Y.; Huang, M.; Duan, N.; and Chen, W. 2023.
\newblock Enhancing retrieval-augmented large language models with iterative retrieval-generation synergy.
\newblock \emph{arXiv preprint arXiv:2305.15294}.

\bibitem[{Shen, Ma, and Wang(2018)}]{shen2018web}
Shen, Z.; Ma, H.; and Wang, K. 2018.
\newblock A web-scale system for scientific knowledge exploration.
\newblock \emph{arXiv preprint arXiv:1805.12216}.

\bibitem[{Sinha et~al.(2015)Sinha, Shen, Song, Ma, Eide, Hsu, and Wang}]{sinha2015overview}
Sinha, A.; Shen, Z.; Song, Y.; Ma, H.; Eide, D.; Hsu, B.-J.; and Wang, K. 2015.
\newblock An overview of microsoft academic service (mas) and applications.
\newblock In \emph{Proceedings of the 24th international conference on world wide web}, 243--246.

\bibitem[{Van~Eck and Waltman(2024)}]{van2024open}
Van~Eck, N.~J.; and Waltman, L. 2024.
\newblock An open approach for classifying research publications.
\newblock \emph{Leiden Madtrics}.

\bibitem[{W3C(2014)}]{schemaorg}
W3C. 2014.
\newblock Schema.org.
\newblock Website.
\newblock Accessed: 2023-11-24.

\bibitem[{{Wikidata contributors}(2024)}]{wikidata}
{Wikidata contributors}. 2024.
\newblock {Wikidata: A free and open knowledge base}.
\newblock \url{https://www.wikidata.org}.
\newblock Accessed: 2024-11-24.

\bibitem[{Zhao et~al.(2023)Zhao, Zhou, Li, Tang, Wang, Hou, Min, Zhang, Zhang, Dong et~al.}]{zhao2023survey}
Zhao, W.~X.; Zhou, K.; Li, J.; Tang, T.; Wang, X.; Hou, Y.; Min, Y.; Zhang, B.; Zhang, J.; Dong, Z.; et~al. 2023.
\newblock A survey of large language models.
\newblock \emph{arXiv preprint arXiv:2303.18223}.

\bibitem[{Zhuang et~al.(2024)Zhuang, Zhang, Cheng, Yang, Liu, Huang, Lin, Rajmohan, Zhang, and Zhang}]{zhuang2024efficientrag}
Zhuang, Z.; Zhang, Z.; Cheng, S.; Yang, F.; Liu, J.; Huang, S.; Lin, Q.; Rajmohan, S.; Zhang, D.; and Zhang, Q. 2024.
\newblock EfficientRAG: Efficient Retriever for Multi-Hop Question Answering.
\newblock \emph{arXiv preprint arXiv:2408.04259}.

\end{thebibliography}

\end{document}